%% file: conference_041818.tex
\documentclass[times, 10pt,twocolumn]{article}
\usepackage{latex8}
\usepackage{times}

\usepackage{graphicx}
\usepackage{amsmath,amssymb,amsfonts}

\title{Distributed Layer-Partitioned Training for Privacy-Preserved Deep Learning}

\author{Chun-Hsien Yu\\
HTC Research \\ Jimmy\_Yu@htc.com \\
\and
Chun-Nan Chou \\
HTC Research \\ jason.cn\_chou@htc.com \\
\and
Emily Chang\\
HTC Research \\ emilyjchang30@gmail.com
}

\begin{document}
\maketitle
\thispagestyle{empty}

\begin{abstract}
Deep Learning techniques have achieved remarkable results in many domains.
Often, training deep learning models requires large datasets, which may require sensitive information to be uploaded to the cloud to accelerate training.
To adequately protect sensitive information, we propose distributed layer-partitioned training with step-wise activation functions for privacy-preserving deep learning.
Experimental results attest our method to be simple and effective.
\end{abstract}

\input{Introduction}
\input{Relatedwork}
\input{DeepLearning}
\input{LayerPartitonLearning}
\input{Experiments}
\input{Conclusion}

\bibliographystyle{latex8}

\end{document}

%% file: Introduction.tex
\section{Introduction}

Since 2012, neural networks and deep architectures have proven very effective in application areas such as computer vision~\cite{he2016deep} and disease diagnosis~\cite{chang2017artificial}.
Although deep architectures such as convolutional neural networks (CNNs)~\cite{fukushima1982neocognitron} emerged as early as the 1980s, they have not risen into the spotlight until now. 
Their recent popularity is due to the rising importance of \emph{scale}, in both data volume and computation resources~\cite{chang2011foundations}.

When the data scale or data volume is small, a local computer can handle the training workload. 
However, when data volume is large, even one-hundredth of AlphaGo's scale, most local computing facilities do not have sufficient computation resources to complete training tasks in a timely fashion.
Take training AlexNet \cite{krizhevsky2012imagenet} as an example.
The task of training one million images took more than one week on a typical local host.
Considering that training parameters such as learning rate and momentum must be experimented upon to find the best settings, tens of rounds of training are required.
Thus, a training task can demand weeks on a typical local host to complete.
This long latency is not feasible for many urgent applications such as national security, investment decisions, and disease outbreak prediction.
The logical solution is to upload data onto the cloud to take advantage of the distributed computing resources.
However, many institutions such as the government, hospitals, and financial institutes forbid data from leaving their local sites for privacy and security reasons.
In this work, we propose a distributed layer-partitioned training methodology that permits metadata (not the original data) to leave a local site, while at the same time preserving data privacy.
The principal idea is that the original data is {\em processed} into {\em irreversible secured metadata} before leaving a local site.
Our method permits substantial training to be run distributively on the {\em irreversible secured metadata} of remote sites, whereas the original data stays at the local secured site.

%% file: Relatedwork.tex
\section{Related Work}

While deep learning is flourishing in many domains, its privacy concerns have attracted much recent attention.
Several pieces of research have shown that training data can be recovered with the access to trained models, which runs the risk of inadvertently revealing sensitive information.
For example, \cite{fredrikson2015model} exploited a model inversion attack to retrieve recognizable images from a facial-recognition model.

Some previous works have attempted to address deep learning privacy concerns. 
\cite{abadi2016deep} adopted differential privacy in the deep learning training process. 
Specifically, they added noises to gradients when applying stochastic gradient descent (SGD).
\cite{shokri2015privacy} devised distributed selective SGD, thereby enabling each client to train its model locally and to exchange model parameters selectively with the global one in the cloud server. 

Although these approaches can introduce certain difficulties in precisely recovering training samples, they are still vulnerable to some attacks such as generative adversarial network-based attacks \cite{hitaj2017deep}.
In contrast, deep learning models trained using our proposed training methodology avoid such attacks since the trained models are separated into two distinct parts.
Some works such as \cite{li2017multi} applied cryptographic techniques to address the privacy issues of deep learning in cloud computing, which is another alternative but often forces learning over encrypted domains.

%% file: DeepLearning.tex
\section{Brief Introduction of CNN}



Convolutional Neural Network (CNN) is a representative model and used by our work to study privacy-preserving schemes.
Throughout this paper, we focus on discussing CNNs, since inputs to CNNs are often images that can be visualized and are easier to illustrate.
However, our proposed training methodology can be applied to all neural networks that have activation functions.
CNNs are typically composed of four computation operations:

\begin{itemize}
\item {\em Matrix multiplication}. In fully connected layers, the computation is the multiplication of two matrices.
\item {\em Convolution}. Convolutional layers apply some designed filters to extract features from the input matrix.
\item {\em Pooling}
In such layers, a function such as {\em max} or {\em average} is used to aggregate data.
\item {\em Non-linear activation}.
In order to model nonlinearity, the neural network introduces the activation function during the evaluation of neuron outputs.
The traditional way to evaluate a neuron output $f$ as a function $g$ of its input $z$ is with $f = g(z)$ where $g$ can be a sigmoid function or a hyperbolic tangent function.
Both of these functions are saturating non-linear.
That is, the range of either function is fixed between a minimum and maximum.
Non-saturating activation function ReLU proposed by \cite{nair2010rectified} can also be used. 
\end{itemize}

Out of the four steps, the first two steps are reversible, in that, if one gets a hold of the output of a neural network layer and its model parameters, one can recreate the layer's input.
The third step is irreversible since the pooling step aggregates inputs using an operator such as average, max, and min.
For example, given an average of $n$ numbers, one cannot recreate the original $n$ numbers.
For the last step, sigmoid and hyperbolic tangent functions are one-to-one functions so both are reversible.
On the other hand, ReLU is an irreversible function, but our experiments showed that inputs to ReLU can be recovered from its outputs well.

%% file: LayerPartitonLearning.tex
\section{Distributed Layer-Partitioned Training}

In this section, we first describe the system architecture of our distributed layer-partitioned training.
We then explain how to derive raw training data out of CNNs' first convolutional layer from its outputs.
Next, we explain how to reverse-engineer the three conventional activation functions.
Finally, we propose a step-wise activation function.

\begin{figure}[t]
  \centering
  \includegraphics[width=0.8\linewidth]{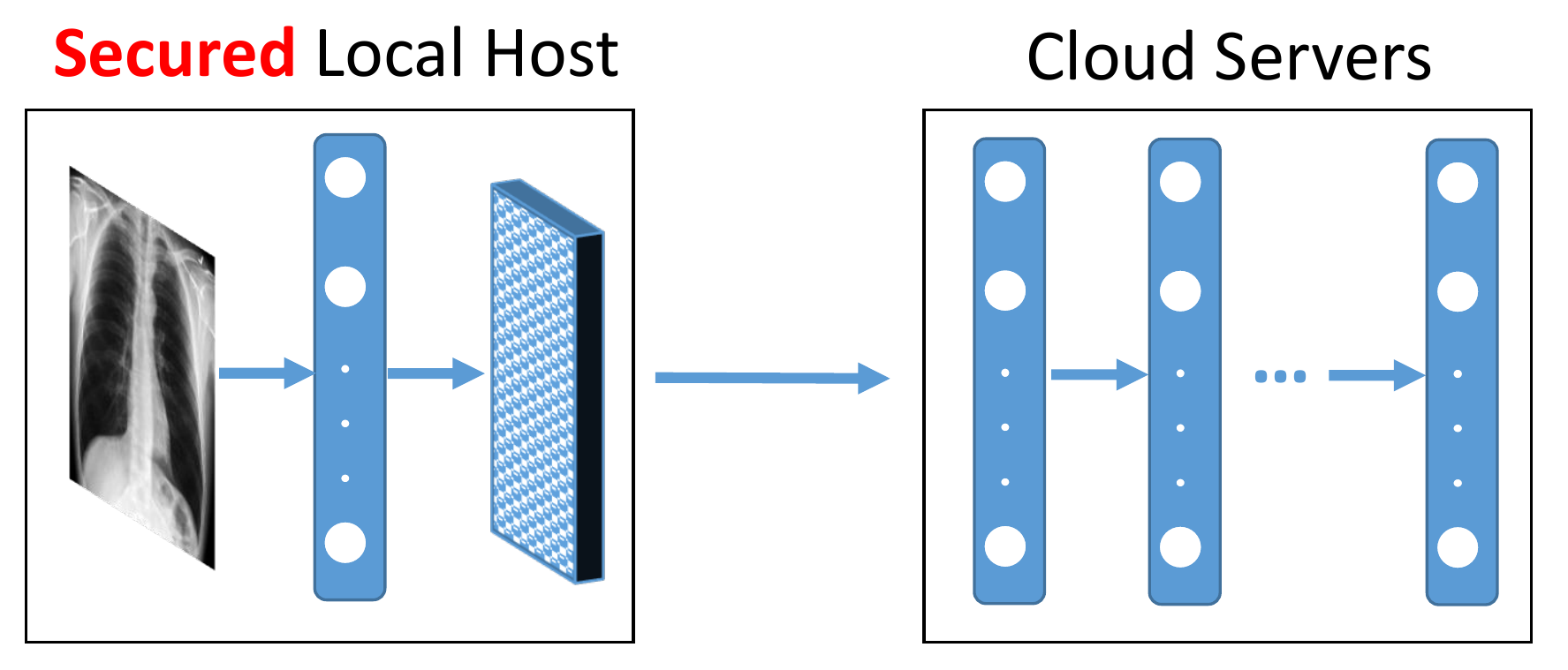}
  \caption{ The high-level view of our system architecture includes: 1) a secured local host having raw data and the first convolutional layer of a CNN model and 2) a cloud server that is responsible for the rest computation. }
  \label{fig:relation}
  \vspace{-.15in}
\end{figure}

\subsection{System Architecture}

Fig.~\ref{fig:relation} illustrates the architecture of our proposed training system.
Our system keeps the design of a CNN model unchanged but partitions it into two different parts.
One is the first convolutional layer that is secured in a local machine and produces the metadata; the other is the rest of the layers that are to be trained on the metadata on distributed sites.

Based on this training architecture, we avoid exposing the raw data out of a local secured site. 
However, if we simply enforce the first convolutional layer of a CNN on the local site, the potential risk of recovering the original data still remains, which we explain in the next subsection. 


\subsection{ Risk of Generic Convolutional Layers }
Assume the input data has the dimension \( ( C_{ \text{in} }, H_{ \text{in} }, W_{ \text{in} } ) \), and the output result and the kernels of the first convolutional layer has the dimension \( ( C_{ \text{out} }, H_{ \text{out} }, W_{ \text{out} } ) \) and the size \( (S \times S) \), respectively.
Thus, the weights of the convolutional kernels in the first layer has the dimension \( ( C_{ \text{out} }, C_{ \text{in} }, S, S) \).
Based on our assumptions, the problem of recovering the input data of the convolutional layer from its output can be formulated as solving a system of linear equations: \( A \boldsymbol{ x } = \boldsymbol{ b } \), where \( \boldsymbol{ x } \) is the input data, \( A \) stands for the weights of the convolutional kernels in the first layer, and \( \boldsymbol{ b } \) is the output of the first layer.
In the linear equations, the number of the unknown parameters we need to solve is \(  C_{ \text{in} } \times H_{ \text{in} } \times W_{ \text{in} }  \), and the number of the equations is \(  C_{ \text{out} } \times H_{ \text{out} } \times W_{ \text{out} }  \).
Therefore, if \( C_{ \text{out} } \) is large enough, this system of linear equations has a unique solution \( x^{ * } \) that is the same as the original input data.
Please note that due to the nature of the convolutional operation, these linear equations are not dependent.

However, if the dimension of the input data is high, e.g. the images in CIFAR-10, the system of linear equations becomes too large to be solved due to the limitation of the computation power. 
Thus, we divide the linear system into many small sub-systems \( \{ A_{i} \boldsymbol{x}_{i} = \boldsymbol{b}_{i} \mid  \forall i \} \), and the solutions \( \{ \boldsymbol{x}_{i}^{*} \mid \forall i \} \) is equal to \( \boldsymbol{x}^{*} \).
In some cases, this may lead to the following situation: \( \boldsymbol{x}^{*} \) is unique, but \( \{  \boldsymbol{x}_{i}^{*} \mid \forall i\} \) is not.
However, if the number of the convolutional kernels is less than or equal to \( C_{\text{out}} \), we can show that
\[
   \boldsymbol{x}^{*} \text{ is unique iff } \{ \boldsymbol{x}_{i}^{*} \mid \forall i \} \text{ is unique.}
\]
In the following, we elaborate the method dividing the original system of linear equations into sub-systems.
First, we denote the weights of the $i^{th}$ convolutional kernel as \( \boldsymbol{\omega}_{i} \) that is a tensor of weights.
After we fix a pair \( (h_{\text{out}}, w_{\text{out}})\), which corresponds to the location of the output results, we can trace back the location of the input tensor, denoted as \( (\mathcal{H}_{\text{in}}, \mathcal{W}_{\text{in}}) \). 
Moreover, this corresponding space \( [\text{Images}]_{(C_{\text{in}}, \mathcal{H}_{\text{in}}, \mathcal{W}_{\text{in}} )} \) is equal to the size of \( \boldsymbol{\omega}_{i} \), for \(i = 1, \ldots, C_{\text{out}}\). 
Therefore, we can state the linear system as 
\[
\begin{split}
  \sum \boldsymbol{\omega}_{1} \odot [\text{Image}]_{(C_{\text{in}}, \mathcal{H}_{\text{in}}, \mathcal{W}_{\text{in}} )} &= [\text{Output}]_{(1, h_{\text{out}}, w_{\text{out}} )} \\
  \vdots & \\
  \sum \boldsymbol{\omega}_{C_{\text{out}}} \odot [\text{Image}]_{(C_{\text{in}}, \mathcal{H}_{\text{in}}, \mathcal{W}_{\text{in}} )} &= [\text{Output}]_{(C_{\text{out}}, h_{\text{out}}, w_{\text{out}} )}, \\
\end{split}
\]
where \( \odot \) is the element-wised products.
It is clear that we have the unique solution \( [\text{Image}]_{(C_{\text{in}}, \mathcal{H}_{\text{in}}, \mathcal{W}_{\text{in}} )}^{*} \) if and only if \( C_{\text{out}} \times \mathcal{H}_{\text{out}} \times \mathcal{W}_{\text{out}} \) is larger than \( C_{\text{in}} \times \mathcal{H}_{\text{in}} \times \mathcal{W}_{\text{in}} \).
Based on the above-mentioned method, there exists an effective way to derive the original input data from the outputs of generic convolutional layers. 
Consequently, if the metadata is caught and the local convoluational layer is compromised, the original data is vulnerable to be leaked.

\subsection{Risk of Conventional Activation Functions}
In this subsection, we explain the approach to derive the inputs of three conventional activation functions from their outputs, revealing the potential privacy problem.
Basically, if we can infer the inverse functions of these three activation functions, the input data can be obtained easily.

The sigmoid function is one of the widely used activation functions. 
Because sigmoid is a bijective function, its unique inverse function exists, namely 
\[ z = \text{sigmoid}^{-1}(y) = -\ln{ ( \frac{ 1 }{ y } - 1 ) }. \]

Similarly, the hyperbolic tangent function is also bijective, and its inverse function is defined as follows:
\[ \text{ tanh }^{-1}(z) = \frac{ \ln{(1+z)} - \ln{(1-z)}}{ 2 }. \]

Unlike the prior two activation functions, ReLU is the only surjective function, which is $ \text{ReLU}(z) = max(0,z)$. 
Thus, we could reverse the result of ReLU only if the output is positive and use the following method to reconstruct the input information. 
First, we pick those equations with positive values.
Second, if the number of linear equations is still insufficient to solve the inputs, we meet the requirement of the equation number by including some of the zero-valued equations.
Lastly, we can solve the system of linear equations as if ReLU is an identity.
	
\subsection{Step-Wise Activation Function} 

\begin{figure}
  \centering
  \includegraphics[width=\linewidth]{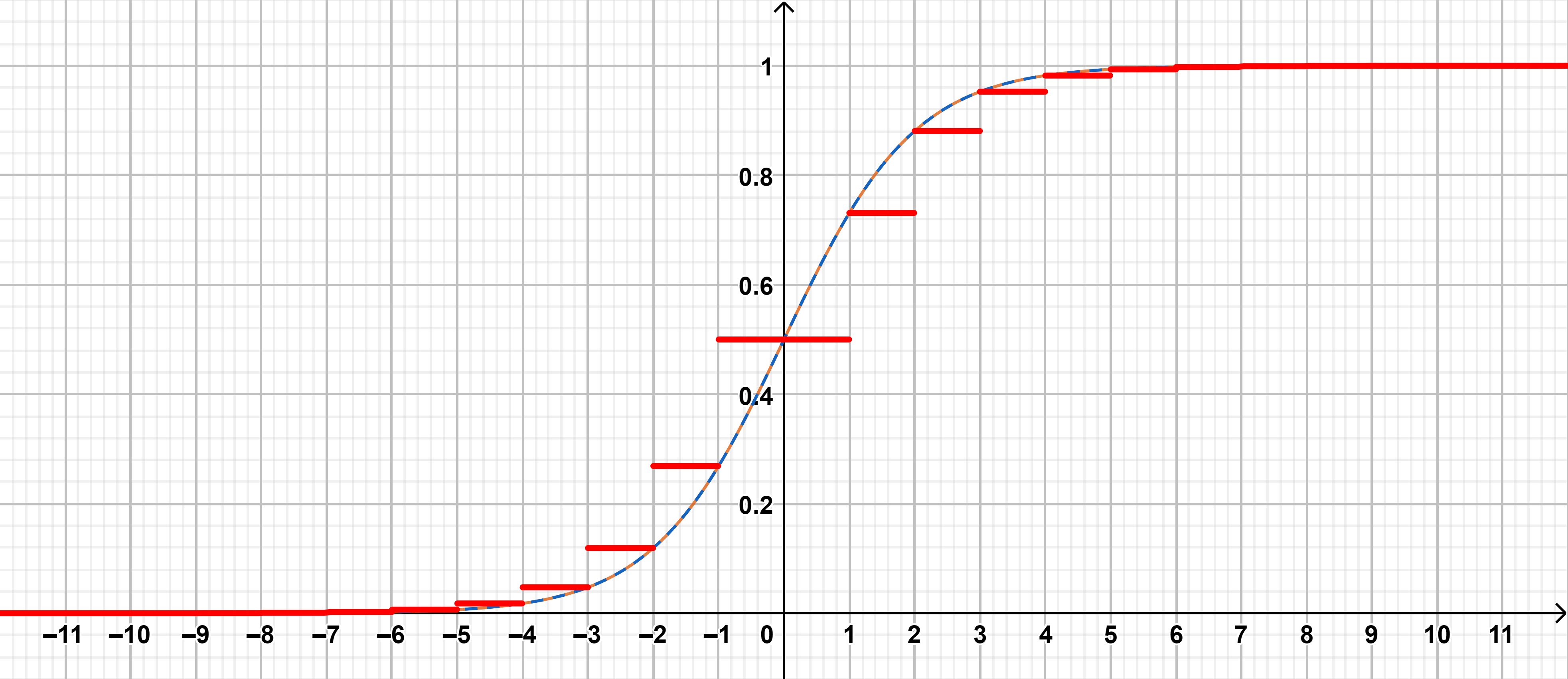}
  \caption{A step-wise activation example.}
  \label{fig:f}
\end{figure}

To make the information irreversible, we modify the conventional activation functions to be step-wise.
We perform quantization on the original functions. 
Fig.~\ref{fig:f} shows one step-wise version of sigmoid function.
Formally, we divide the input domain into intervals and make the output step-wise.
Given the number of intervals $n$, the clipping value $v$ and an activation function $g$, the step-wise version of $g$ is  
\[ g^{\text{step}}(x) = g( \text{sign}(x) \cdot \lfloor\frac{\min (|x|, v)}{v/n}\rfloor \cdot \frac{v}{n}), \]
where $\text{sign}(\cdot)$ is a function that returns $1$ if its input is non-negative, and $-1$ otherwise.

%% file: Experiments.tex
\section{Preliminary Evaluation}

We performed experiments to verify that 1) models trained using our proposed method can reach equivalent accuracy, and 2) raw images cannot be reversed after our proposed step-wise functions were applied. 
We applied our method to MobileNetV2~\cite{sandler2018mobilenetv2} on two standard image datasets: MNIST \cite{lecun1998gradient} and CIFAR-10 \cite{krizhevsky2009learning}.
We chose these two public datasets because they have a long record of serving as benchmarks in machine learning \cite{abadi2016deep}.
In all of our experiments, we utilized SGD with momentum to train the models and ran $90$ epochs by using the same initialization method and hyper-parameter settings.

To examine the effect of our method on prediction accuracy, we substituted our step-wise sigmoid functions with different interval values $n$ for the first activation function in MobileNetV2, whereas the other components remain unchanged. 
Table.~\ref{table:exp_accuracy} shows that the accuracy of the model using our step-wise sigmoid increases as the value of $n$ increases.
In comparison with the accuracy of the model using the original sigmoid, the accuracy degradation of using our step-wise sigmoid with $n=21$ is negligible for MNIST.
This result provides us a setting to maintain equivalent accuracy.
On the more complicated CIFAR-10 \cite{krizhevsky2009learning} dataset, the accuracy degradation using our privacy-preserved is less competitive.
A further increase in the interval value $n$ is expected to increase its accuracy.  

\begin{figure}[t]
  \centering
  \includegraphics[width=0.8\linewidth]{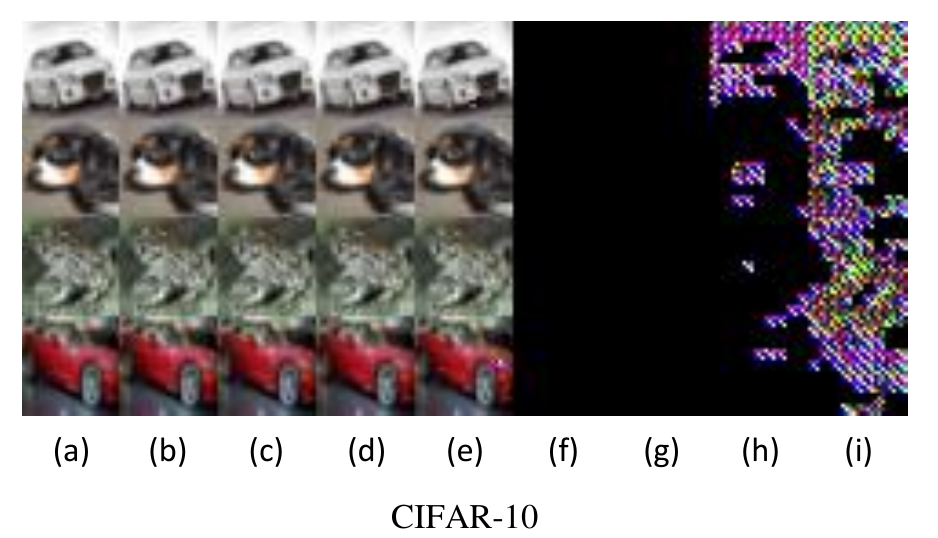}
   \includegraphics[width=0.8\linewidth]{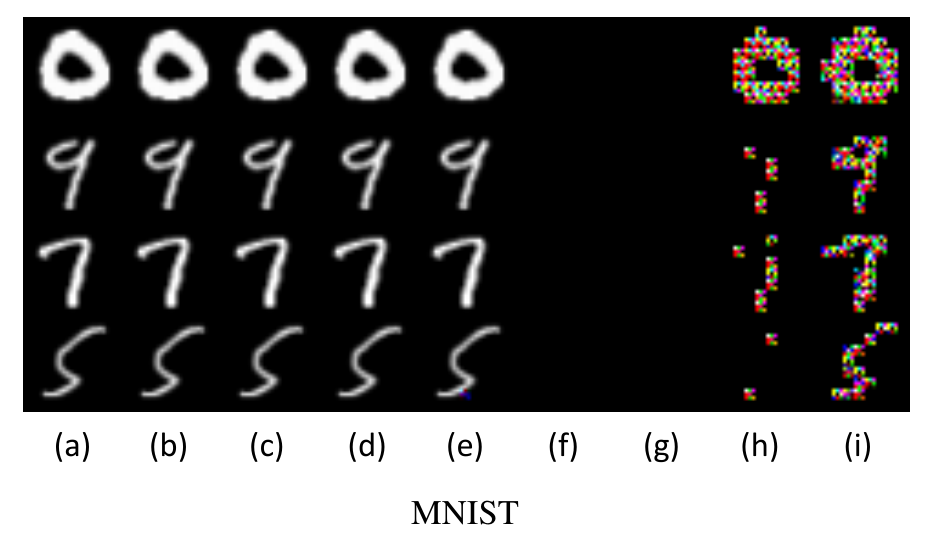}
   \vspace{-.1in}
  \caption{
  (a) original images, (b) $\sim$ (e) reversed results of convolutional layers w/o activation function, w/ sigmoid, w/ hyperbolic tangent, w/ ReLU, (f) $\sim$ (i) reversed results of our proposed step-wise function on sigmoid with the clipping value $v=10$ and the number of intervals $n$ set to 3, 5, 11 and 21.
  }
  \label{fig:experiment}
  \vspace{-.2in}
\end{figure}

To evaluate the capability of preserving data privacy, we visualized the input images from the outputs of the first convolutional layer in our experimental models by using the method described in the previous section.
The results shown in Fig.~\ref{fig:experiment} demonstrate that most reversed inputs of our proposed step-wise sigmoid functions are unrecognizable.
Qualitatively, this result demonstrates that our proposed scheme to be effective in preserving data privacy.

\begin{table}[tbp]
\centering
\caption{Accuracy with and without step-wise activation function. M and C stand for MINST and CIFAR-10, respectively.}
\begin{tabular}{cccccc}
\hline
\textit{} & \textbf{Sig.} & \textbf{\begin{tabular}[c]{@{}c@{}}Step.\\ n = 3\end{tabular}} & \textbf{\begin{tabular}[c]{@{}c@{}}Step.\\ n = 5\end{tabular}} & \textbf{\begin{tabular}[c]{@{}c@{}}Step.\\ n = 11\end{tabular}} & \textbf{\begin{tabular}[c]{@{}c@{}}Step.\\ n = 21\end{tabular}} \\ \hline
\textbf{M.} & 99.68\% & 10.28\% & 23.27\% & 99.57\% & 99.65\% \\
\textbf{C.} & 86.94\% & 13.74\% & 23.45\% & 49.91\% & 81.28\% \\ \hline
\end{tabular}
\label{table:exp_accuracy}
\end{table}

%% file: Conclusion.tex
\section{Conclusion}

Our proposed distributed layer-partition training with step-wise activation functions can protect the original data against malicious attacks better. 
The preliminary evaluation shows our proposed methodology successfully creates considerable difficulties for recovering raw data even if the metadata is captured and the underlying weights are compromised by adversaries.
We also observe that tradeoffs exist between accuracy and data privacy.
Based on our observation, a hospital can set the interval parameter properly to achieve the desired balance between accuracy and privacy.
As future work, we will investigate what the ideal interval values of different settings are and extend our idea to work with other tasks such as object detection and segmentation.